\documentclass[lettersize,journal]{IEEEtran}

\usepackage{textcomp}
\usepackage{stfloats}
\usepackage{microtype}
\usepackage{xcolor}
\usepackage{lmodern}

\usepackage{amsmath}
\usepackage{amsfonts}
\usepackage{amssymb}
\usepackage{mathtools}
\usepackage{amsthm}
\usepackage{stmaryrd}
\usepackage{nicefrac}
\DeclareMathAlphabet{\mymathbb}{U}{BOONDOX-ds}{m}{n}

\usepackage{graphicx}
\usepackage{array}
\usepackage{booktabs}
\usepackage[caption=false,font=footnotesize]{subfig}
\usepackage{wrapfig}
\usepackage{adjustbox}
\usepackage{tikz}
\usetikzlibrary{shapes.geometric, positioning, fit}

\usepackage{verbatim}
\usepackage{comment}

\usepackage{cite}
\usepackage{url}
\usepackage{hyperref}
\usepackage[capitalize,noabbrev]{cleveref}

\usepackage{amsmath,amsfonts,bm}

\def\va{{\bm{a}}}

\def\vd{{\bm{d}}}

\def\vg{{\bm{g}}}

\def\vi{{\bm{i}}}
\def\vj{{\bm{j}}}
\def\vk{{\bm{k}}}

\def\vp{{\bm{p}}}

\def\vr{{\bm{r}}}

\def\vx{{\bm{x}}}
\def\vy{{\bm{y}}}

\def\evg{{g}}

\def\evi{{i}}

\def\evx{{x}}

\def\mA{{\bm{A}}}
\def\mB{{\bm{B}}}

\def\mH{{\bm{H}}}
\def\mI{{\bm{I}}}

\def\mT{{\bm{T}}}

\def\mX{{\bm{X}}}

\DeclareMathAlphabet{\mathsfit}{\encodingdefault}{\sfdefault}{m}{sl}
\SetMathAlphabet{\mathsfit}{bold}{\encodingdefault}{\sfdefault}{bx}{n}
\newcommand{\tens}[1]{\bm{\mathsfit{#1}}}
\def\tA{{\tens{A}}}

\def\tG{{\tens{G}}}

\def\tI{{\tens{I}}}

\def\tM{{\tens{M}}}

\def\tT{{\tens{T}}}

\def\sR{{\mathbb{R}}}

\def\emA{{A}}
\def\emB{{B}}

\def\emX{{X}}

\newcommand{\etens}[1]{\mathsfit{#1}}

\def\etA{{\etens{A}}}

\def\etG{{\etens{G}}}

\def\etM{{\etens{M}}}

\def\etO{{\etens{O}}}

\def\etR{{\etens{R}}}

\def\etT{{\etens{T}}}

\usetikzlibrary{positioning,fit,shapes.geometric,calc}

\colorlet{colorMPS}{orange!20}       %
\colorlet{colorMPO}{green!20}        %
\colorlet{colorH}{blue!20}           %
\colorlet{colorD}{yellow!20}            %
\colorlet{colorInput}{red!10}     %
\colorlet{colorBlock}{gray!10}       %

\def\nodesize{7mm}                   %
\def\inputsize{5mm}                  %
\def\leglength{0.3cm}                %
\def\legsep{0.2cm}                   %
\def\textsep{0.1cm}                  %
\def\blockheight{7mm}                %
\def\layersep{1.0cm}                 %

\tikzset{
  indexlabel/.style={
    font=\tiny,
    minimum width=0.4cm,
    inner sep=0pt,
    anchor=east
  },
  tensor/.style={
    draw,
    circle,
    minimum size=\nodesize,
    inner sep=0pt,
    fill=colorMPS
  },
  mpo/.style={
    draw,
    rectangle,
    minimum size=\nodesize,
    inner sep=0pt,
    fill=colorMPO
  },
  mponofill/.style={
    draw,
    rectangle,
    minimum size=\nodesize,
    inner sep=0pt,
    fill=colorMPO
  },
  input/.style={
    draw,
    circle,
    minimum size=\inputsize,
    inner sep=0pt,
    fill=colorInput
  },
  inputwide/.style={
    draw,
    rounded corners=2pt,
    rectangle,
    minimum height=\inputsize,
    minimum width=\nodesize,
    inner sep=2pt,
    fill=colorInput
  },
  itensor/.style={
    draw,
    rectangle,
    minimum size=\nodesize,
    inner sep=0pt,
    fill=colorH
  },
  gtensor/.style={
    draw,
    rectangle,
    minimum size=\nodesize,
    inner sep=0pt,
    fill=colorBlock
  },
  htensor/.style={
    draw,
    rectangle,
    minimum size=\nodesize,
    inner sep=0pt,
    fill=colorH
  },
  dtensor/.style={
    draw,
    diamond,
    minimum size=\nodesize,
    inner sep=0pt,
    fill=colorD
  },
  block/.style={
    draw,
    rectangle,
    minimum height=\blockheight,
    fill=colorBlock
  }
}

\newcommand{\fivedtensor}{
  \begin{tikzpicture}[font=\small]
    \node[tensor] (T) {$\etT$};
    \draw (T.north) -- node[right=0cm, font=\tiny] {$i$} ++(0,\leglength);
    \draw (T.south) -- node[left=0cm, font=\tiny] {$j$} ++(0,-\leglength);
    \draw (T.west) -- node[above=0cm, font=\tiny] {$k$} ++(-\leglength,0);
    \draw (T.east) -- node[below=0cm, font=\tiny] {$l$} ++(\leglength,0);
    \draw (T.45) -- node[right=\textsep, font=\tiny] {$m$} ++(0.5*\leglength,0.5*\leglength);
  \end{tikzpicture}
}

\newcommand{\matrixproduct}{
  \begin{tikzpicture}[font=\small]
    \node[tensor] (A) {$A$};
    \node[tensor, right=\legsep of A] (B) {$B$};
    \draw (A.west) -- node[above=0cm, font=\tiny] {$i$} ++(-\leglength,0);
    \draw (A.east) -- (B.west) node[midway, below=\textsep, font=\tiny] {$j$};
    \draw (B.east) -- node[above=0cm, font=\tiny] {$k$} ++(\leglength,0);
  \end{tikzpicture}
}

\newcommand{\mpstypeII}{
  \begin{tikzpicture}[font=\small]
    \node[tensor] (A1) at (0,0) {$\tT_1$};
    \node[tensor] (A2) at (1,0) {$\tT_2$};
    \node (Adots) at (2,0) {$\cdots$};
    \node[tensor] (AN) at (3,0) {$\tT_N$};
    \node[input] (x1) at (0,-\layersep) {$x$};
    \node[input] (x2) at (1,-\layersep) {$x$};
    \node[input] (xN) at (3,-\layersep) {$x$};
    \draw (A1.east) -- (A2.west);
    \draw (A2.east) -- (Adots.west);
    \draw (Adots.east) -- (AN.west);
    \foreach \i/\N/\X in {1/A1/x1,2/A2/x2,N/AN/xN}{
      \draw (\N.south) -- (\X.north);
      \node[indexlabel] at ([xshift=-0.15cm, yshift=-0.2cm]\N.south) {$d_{\i}$};
    }
    \draw (AN.north) -- ++(\legsep,\legsep);
    \node[indexlabel, anchor=west] at ([xshift=0.35cm, yshift=0.0cm]AN.north) {out};
  \end{tikzpicture}
}

\newcommand{\mposqtypeII}{
  \begin{tikzpicture}[font=\small]
    \node[tensor] (T1) at (0,0) {$\tT_1$};
    \node[tensor] (T2) at (1,0) {$\tT_2$};
    \node (Tdots) at (2,0) {$\cdots$};
    \node[tensor] (TN) at (3,0) {$\tT_N$};

    \node[mpo] (A1) at (0,-\layersep) {$\tA_1$};
    \node[mpo] (A2) at (1,-\layersep) {$\tA_2$};
    \node (Adots) at (2,-\layersep) {$\cdots$};
    \node[mpo] (AN) at (3,-\layersep) {$\tA_N$};

    \node[input] (x1) at (0,-2*\layersep) {$x$};
    \node[input] (x2) at (1,-2*\layersep) {$x$};
    \node[input] (xN) at (3,-2*\layersep) {$x$};

    \draw (T1.east) -- (T2.west);
    \draw (T2.east) -- (Tdots.west);
    \draw (Tdots.east) -- (TN.west);

    \draw (A1.east) -- (A2.west);
    \draw (A2.east) -- (Adots.west);
    \draw (Adots.east) -- (AN.west);

    \foreach \i/\T/\A in {1/T1/A1,2/T2/A2,N/TN/AN}{
      \draw (\T.south) -- (\A.north);
    }

    \foreach \i/\A/\X in {1/A1/x1,2/A2/x2,N/AN/xN}{
      \draw (\A.south) -- (\X.north);
      \node[indexlabel] at ([xshift=-0.15cm, yshift=-0.2cm]\A.south) {$d_{\i}$};
    }
    \draw (TN.north) -- ++(\legsep,\legsep);
    \node[indexlabel, anchor=west] at ([xshift=0.35cm, yshift=0.0cm]TN.north) {out};
  \end{tikzpicture}
}

\newcommand{\cpd}{
  \begin{tikzpicture}[font=\small]
    \node[tensor] (T1) at (0,0) {$\tT_1$};
    \node[tensor] (T2) at (1,0) {$\tT_2$};
    \node (Tdots) at (2,0) {$\cdots$};
    \node[tensor] (TN) at (3,0) {$\tT_N$};

    \node[block, fit=(T1.north west)(TN.north east), anchor=south, yshift=\layersep/2, inner xsep=2pt] (G) {$\mathcal{G}$};

    \node[input] (x1) at (0,-\layersep) {$x$};
    \node[input] (x2) at (1,-\layersep) {$x$};
    \node[input] (xN) at (3,-\layersep) {$x$};

    \foreach \i/\T in {1/T1,2/T2,N/TN}{
      \draw (\T.north) -- (G.south -| \T.north);
      \node[indexlabel] at ([xshift=-0.15cm, yshift=0.15cm]\T.north) {$r_{\i}$};
    }

    \foreach \i/\T/\X in {1/T1/x1,2/T2/x2,N/TN/xN}{
      \draw (\T.south) -- (\X.north);
      \node[indexlabel] at ([xshift=-0.15cm, yshift=-0.2cm]\T.south) {$d_{\i}$};
    }
  \end{tikzpicture}
}

\newcommand{\mpo}{
  \begin{tikzpicture}
    \node[mponofill] (O1) {$\etO_1$};
    \node[mponofill, right=\legsep of O1] (O2) {$\etO_2$};
    \node[right=\legsep of O2] (dots2) {$\cdots$};
    \node[mponofill, right=\legsep of dots2] (Om) {$\etO_n$};

    \draw (O1.north) -- node[left=0mm, yshift=1mm, font=\tiny] {$i'_1$} ++(0,\leglength);
    \draw (O2.north) -- node[left=0mm, yshift=1mm, font=\tiny] {$i'_2$} ++(0,\leglength);
    \draw (Om.north) -- node[left=0mm, yshift=1mm, font=\tiny] {$i'_n$} ++(0,\leglength);

    \draw (O1.south) -- node[left=0mm, yshift=-1mm, font=\tiny] {$i_1$} ++(0,-\leglength);
    \draw (O2.south) -- node[left=0mm, yshift=-1mm, font=\tiny] {$i_2$} ++(0,-\leglength);
    \draw (Om.south) -- node[left=0mm, yshift=-1mm, font=\tiny] {$i_n$} ++(0,-\leglength);
    \draw (O1.east) -- (O2.west);
    \draw (O2.east) -- (dots2.west);
    \draw (dots2.east) -- (Om.west);
  \end{tikzpicture}
}

\newcommand{\cumsummpo}{
  \begin{tikzpicture}[font=\small]
    \node[tensor] (A1) at (0,0) {$\tT_1$};
    \node[tensor] (A2) at (2.0,0) {$\tT_2$};
    \node (Adots) at (3.0,0) {$\cdots$};
    \node[tensor] (AN) at (4.8,0) {$\tT_N$};

    \node[dtensor] (D1) at (0,-\layersep) {$\mathcal{I}$};
    \node[htensor] (H0) at (-1.0,-\layersep) {$\bm{\Theta}$};
    \node[htensor] (H1) at (1.0,-\layersep) {$\bm{\Theta}$};
    \node[dtensor] (D2) at (2.0,-\layersep) {$\mathcal{I}$};
    \node (HDdots) at (3.0,-\layersep) {$\cdots$};
    \node[htensor] (H2) at (4,-\layersep) {$\bm{\Theta}$};
    \node[dtensor] (DN) at (4.8,-\layersep) {$\mathcal{I}$};

    \draw (A1.east) -- (A2.west);
    \draw (A2.east) -- (Adots.west);
    \draw (Adots.east) -- (AN.west);

    \draw (A1.south) -- (D1.north);
    \draw (A2.south) -- (D2.north);
    \draw (AN.south) -- (DN.north);

    \draw (H0.east) -- (D1.west);
    \draw (D1.east) -- (H1.west);
    \draw (H1.east) -- (D2.west);
    \draw (D2.east) -- (HDdots.west);
    \draw (HDdots.east) -- (H2.west);
    \draw (H2.east) -- (DN.west);

    \draw (D1.south) -- ++(0,-\leglength) node[input, below=\leglength of D1.south] (x1) {$x$};
    \draw (D2.south) -- ++(0,-\leglength) node[input, below=\leglength of D2.south] (x2) {$x$};
    \draw (DN.south) -- ++(0,-\leglength) node[input, below=\leglength of DN.south] (xN) {$x$};
  \end{tikzpicture}
}

\newcommand{\cumsumblock}{
  \begin{tikzpicture}[font=\small]
    \node[htensor] (H) {$\bm{\Theta}$};

    \node[dtensor, right=\leglength of H] (D) {$\mathcal{I}$};

    \draw (H.east) -- (D.west);
    \node[above=0.0cm of H.east, xshift=\legsep, font=\tiny] {$k$};

    \draw (H.west) -- ++(-\leglength,0);
    \node[left=0.1cm of H.west,yshift=0.15cm, font=\tiny] {$a_i$};

    \draw (D.north) -- ++(0,\leglength);
    \node[above=0.1cm of D.north, xshift=0.15cm, font=\tiny] {$i'_i$};
    \draw (D.south) -- ++(0,-\leglength);
    \node[below=0.1cm of D.south, xshift=0.15cm, font=\tiny] {$i_i$};
    \draw (D.east) -- ++(\leglength,0);
    \node[right=0.0cm of D.east, yshift=0.15cm, font=\tiny] {$a_{i+1}$};

    \node[right=0.7cm of D] (equals) {$=$};

    \node[mpo, right=0.7cm of equals] (C) {$\tA$};

    \draw (C.west) -- ++(-\leglength,0);
    \draw (C.east) -- ++(\leglength,0);
    \draw (C.north) -- ++(0,\leglength);
    \draw (C.south) -- ++(0,-\leglength);

    \node[left=0.1cm of C.west, yshift=0.15cm, font=\tiny] {$a_i$};
    \node[right=0.0cm of C.east, yshift=0.15cm, font=\tiny] {$a_{i+1}$};
    \node[above=0.1cm of C.north, xshift=0.15cm, font=\tiny] {$i'_i$};
    \node[below=0.1cm of C.south, xshift=0.15cm, font=\tiny] {$i_i$};

    \node[above=\legsep of equals, font=\small] {$\sum_k$};
  \end{tikzpicture}
}

\newcommand{\cmpotwo}{
    \begin{tikzpicture}[font=\small, baseline=(current bounding box.center)]
      \node[tensor] (T1) at (0,0) {$\tT_1$};
      \node[tensor] (T2) at (1,0) {$\tT_2$};
      \node (Tdots) at (2,0) {$\cdots$};
      \node[tensor] (TN) at (3,0) {$\tT_N$};

      \node[inputwide] (X1) at (0,-\layersep) {$\mX_1$};
      \node[inputwide] (X2) at (1,-\layersep) {$\mX_2$};
      \node (Xdots) at (2,-\layersep) {$\cdots$};
      \node[inputwide] (XN) at (3,-\layersep) {$\mX_N$};

      \node[tensor] (B1) at (0,-2*\layersep) {$\tG_1$};
      \node[tensor] (B2) at (1,-2*\layersep) {$\tG_2$};
      \node (Bdots) at (2,-2*\layersep) {$\cdots$};
      \node[tensor] (BN) at (3,-2*\layersep) {$\tG_N$};

      \draw (T1.east) -- (T2.west);
      \draw (T2.east) -- (Tdots.west);
      \draw (Tdots.east) -- (TN.west);

      \draw (B1.east) -- (B2.west);
      \draw (B2.east) -- (Bdots.west);
      \draw (Bdots.east) -- (BN.west);

      \foreach \i/\T/\X/\B in {1/T1/X1/B1, 2/T2/X2/B2, N/TN/XN/BN}{
        \draw (\T.south) -- (\X.north);
        \draw (\X.south) -- (\B.north);
      }
    \end{tikzpicture}
}

\newcommand{\cmpotwoasmpo}{
  \begin{tikzpicture}[font=\small, baseline=(current bounding box.center)]

    \foreach \ip/\lab in {0/1,2/2,5/N}{
      \node[itensor] (I\ip) at (\ip,-1.05*\layersep) {$\tI\!\otimes\!\tI$};
    }
    \foreach \gp/\lab in {1/1,3/2,6/N}{
      \node[tensor] (G\gp) at (\gp,-1.05*\layersep) {$\tG_{\lab}$};
    }
    \node (IGdots) at (4,-1.05*\layersep) {$\cdots$};

    \draw (I0.east) -- (G1.west);
    \draw (G1.east) -- (I2.west);
    \draw (I2.east) -- (G3.west);
    \draw (G3.east) -- (IGdots.west);
    \draw (IGdots.east) -- (I5.west);
    \draw (I5.east) -- (G6.west);

    \foreach \ip/\lab in {0/1,2/2,5/N}{
      \node[tensor] (T\ip) at (\ip,0) {$\tT_{\lab}$};
      \draw (T\ip.south) -- (I\ip.north);
    }
    \node (Tdots) at (4,0) {$\cdots$};
    \draw (T0.east) -- (T2.west);
    \draw (T2.east) -- (Tdots.west);
    \draw (Tdots.east) -- (T5.west);

    \foreach \ip/\gp/\lab in {0/1/1,2/3/2,5/6/N}{
      \coordinate (Xc\ip) at ($(I\ip.south)!0.5!(G\gp.south)$);
      \node[inputwide, minimum width=1.8*\nodesize] (X\ip)
        at ([yshift=-0.6*\layersep]Xc\ip) {$\mX_{\lab}$};
      \draw (X\ip.north -| I\ip.south) -- (I\ip.south);
      \draw (X\ip.north -| G\gp.south) -- (G\gp.south);
    }
    \node (Xdots) at (4,-1.55*\layersep) {$\cdots$};
  \end{tikzpicture}
}

\newcommand{\cmpotwoasmpoblock}{
  \begin{tikzpicture}[font=\small, baseline=(current bounding box.center),
    isq/.style={itensor, minimum size=6mm},
    gsq/.style={gtensor, minimum size=6mm}]
    \node[isq] (I)      at (0,0)    {$\tI$};                       
    \node      (otimes) at (1,0)    {$\displaystyle\bigotimes$};   
    \node[isq] (K)      at (0,-1.3) {$\tI$};                       
    \node[tensor] (G)      at (1,-1.3) {$\tG$};                    

    \draw (K.east) -- (G.west);

    \node[fit=(I)(otimes)(K)(G), inner sep=4pt] (box) {};

    \draw (I.north) -- ++(0,\leglength);
    \node[above=0.05cm of I.north, xshift=0.18cm, font=\tiny] {$p'_n$};
    \draw (I.south) -- ++(0,-\leglength);
    \node[below=0.05cm of I.south, xshift=0.18cm, font=\tiny] {$p_n$};
    \draw (K.west) -- ++(-\leglength,0);
    \node[left=0.02cm of K.west, yshift=0.15cm, font=\tiny] {$a_n$};
    \draw (G.north) -- ++(0,\leglength);
    \node[above=0.05cm of G.north, xshift=0.18cm, font=\tiny] {$k_n$};
    \draw (G.east) -- ++(\leglength,0);
    \node[right=0.02cm of G.east, yshift=0.15cm, font=\tiny] {$a_{n+1}$};

    \node[right=0.15cm of box] (equals) {$=$};

    \node[mpo, right=0.5cm of equals] (A) {$\tA$};
    \draw (A.west) -- ++(-\leglength,0);
    \draw (A.east) -- ++(\leglength,0);
    \draw (A.north) -- ++(0,\leglength);
    \draw (A.south) -- ++(0,-\leglength);
    \node[left=0.05cm of A.west, yshift=0.15cm, font=\tiny] {$a_n$};
    \node[right=0.0cm of A.east, yshift=0.15cm, font=\tiny] {$a_{n+1}$};
    \node[above=0.2cm of A.north, xshift=0.15cm, font=\tiny] {$(k_n,p'_n)$};
    \node[below=0.1cm of A.south, xshift=0.18cm, font=\tiny] {$p_n$};
  \end{tikzpicture}
}

\newlength{\diagramsep}
\def\Figref#1{Figure~\ref{#1}}
\def\Eqref#1{Equation~\ref{#1}}

\theoremstyle{plain}

\theoremstyle{definition}

\theoremstyle{remark}

\begin{document}

\title{(MPO)²: Multivariate Polynomial Optimization based on Matrix Product Operators}

\author{Niccol\`{o} Ciolli, Anders Vestergaard Nørskov, Michael Kastoryano, Petr Taborsky, Morten Mørup}%

\maketitle
\begin{abstract}
Central to machine learning and signal processing is the ability to perform universal function approximation and learn complex input-output relationships from limited numbers of observations.
Multivariate polynomial models offer a natural way to express such relationships through multiplicative feature interactions, but their coefficient tensors grow exponentially in size with the polynomial degree.
Existing tensorized polynomial models reduce this cost, yet canonical polyadic decompositions have rank-limited expressivity, and tensor train formulations are feature order dependent.
We introduce Multivariate Polynomial Optimization based on Matrix Product Operators (MPO)$^2$, a framework that combines learned MPO feature embeddings with compact polynomial weight tensors.
This yields feature order independent polynomial representations that can incorporate structured operators such as projections, convolutions, and masks for weight tensor symmetries.
Across regression and classification benchmarks, (MPO)$^2$ improves over existing tensor decomposition based polynomial models and provides a flexible alternative for efficient polynomial function approximation.\footnote{An early version of this article has been presented as a spotlight paper, non-archival at the ICML 2026 workshop CoLoRAI26 -- The 2nd Workshop on
Connecting Low-rank
Representations in AI~\cite{ciolli2025mpoWorkshop}}
\end{abstract}


\section{Introduction}
A fundamental objective of machine learning and signal processing is to learn suitable functions from data that can map inputs to associated outputs and generalize to unseen data.
Whereas it is well established that deep learning can provide universal function approximation for sufficiently large model architectures~\cite{hornik1989multilayer}, the models often are challenging to interpret.
Other modeling tools for universal function approximations can overcome the interpretability problem.
This includes Gaussian Processes (GPs) for suitable choices of kernels~\cite{williams2006gaussian,tran2016variational} and series approximations.

In the recent decade context aware learning methods have demonstrated superior performance in generalization while leveraging non-linear dependencies.
This includes the transformer architecture~\cite{vaswani2017attention}, gating mechanisms as used for instance in long-short term memory~\cite{hochreiter1997long} and gated linear units~\cite{dauphin2017language}.
Importantly, such architectures directly operate multiplicative interactions between attributes, instrumental in deep learning~\cite{jayakumar2020multiplicative}.
Classical statistics easily model multiplicative interactions using standard interaction terms.
In contrast, a standard multilayer feedforward network requires at least four hidden neurons just to represent a simple multiplication operation~\cite{lin2017does}.

Conversely, polynomial networks directly express multiplicative interactions using higher degree terms.
Early works on polynomial networks were based on the pi-sigma network~\cite{shin1991pi} that expressed polynomials as multiplications of simple linear regression functions.
Ridge polynomial networks~\cite{shin1995ridge} similarly express the polynomial function in terms of successive accumulations whereas the pi-sigma-pi network introduced an additional multiplicative layer combining multiple pi-sigma networks~\cite{li2003sigma}.
More recently, higher order Sigma–Pi and Sigma–Pi–Sigma neural networks (SPSNNs) were proposed~\cite{jiao2024spsnn, deng2024rspsnn, sarikaya2023training}.
They use multiplicative units in place of neurons and parameter sharing over layers to compactly encode polynomial maps.

Theoretically, multiplicative networks can approximate smooth targets with fewer layers/neurons than ReLU nets~\cite{benshaul2023multiplicative, jayakumar2020multiplicative}, while biologically inspired multiplicative couplings accelerate learning and gating in RNNs~\cite{zhang2025multiplicative}.

Recently, polynomial networks have advanced to exploring tensor decomposition structures including the canonical polyadic decomposition (CPD)~\cite{hendrikx2019algebraic, ayvaz2021eusipco,govindarajan2022regression,Ayvaz-2022-CPD-StructuredMulti,kilic2025interpretablebayesiantensornetwork} as well as hierarchically coupled CPD decompositions forming the $\Pi$-Net~\cite{Chrysos2021,chrysos2022augmenting} which are closely related respectively to the pi-sigma and ridge polynomial networks that can be considered rank one CPD structured.
Besides the CPD, these approaches have also been advanced to more flexible tensor network structures including the tensor train/matrix product states decomposition (TT/MPS)~\cite{stoudenmire2016supervised,götte2021blocksparsetensortrainformat,kilic2025interpretablebayesiantensornetwork}.
Furthermore, tensor Machines learn target–specific polynomial features via low–rank CPD tensors~\cite{yang2015tm}.
These typically assume CP/Tucker parameterizations and squared/logistic losses.
From the input-output mode perspective, they can be seen as a variants of multivariate polynomial models of more recent work~\cite{ayvaz2021eusipco}.

Importantly, decomposed polynomial networks can be optimized using simple alternating linear systems (ALS) optimization using second order methods to optimize each factor of the decomposition at a time, see also~\cite{hendrikx2019algebraic,Ayvaz-2022-CPD-StructuredMulti,kilic2025interpretablebayesiantensornetwork}.
Whereas the above polynomial networks explore tensor decompositions for regression we note that they differ from tensor regression which aims to explore regression of high order data structures~\cite{liu2022tensor}.
Tensor network representation for polynomial networks also differ from recent efforts to use tensor decomposition procedures to compress the weight tensors in deep learning models.
For a discussion of the connections between tensor decompositions and deep learning, see also~\cite{panagakis2021tensor}.

Crucially, tensor decomposition structures address the curse of dimensionality of the multivariate polynomial regression weights~\cite{shin1991pi}.
However, the existing formulations using the CPD decomposition is theoretically bounded in expressive power, whereas the current TT/MPS based modeling procedures~ \cite{stoudenmire2016supervised,efthymiou2019tensornetworkmachinelearning,götte2021blocksparsetensortrainformat,kilic2025interpretablebayesiantensornetwork} are feature order dependent imposing feature specific blocks of the decomposition.
Furthermore, previous procedures do not account for redundancies in the weight tensor and relies on prespecified feature representations.
These limitations, we argue, have hampered the wider adoption of this otherwise attractive alternative to deep learning based function approximation.

We presently propose the Multivariate Polynomial Optimization based on Matrix Product Operators (MPO)$^2$ framework, a new tensor network based structure for the modeling of higher order polynomials.
Notably, (MPO)$^2$ generalizes polynomial tensor networks enhancing:
\begin{itemize}
  \item \textbf{Expressiveness:}
    We consider more expressive tensor network representations exploring the matrix product operators formalism to both learn feature and polynomial representations with added expressive capabilities when compared to CPD and existing MPS/TT based procedures notably also being feature order independent when compared to the latter.
  \item \textbf{Reliability and scalability:}
    We devise an alternating least-squares procedure providing closed-form updates as well as support for scalable gradient descent methods.
  \item \textbf{Versatility:}
    We introduce generic structured operators to account for inductive biases such as polynomial degree redundancies and translation invariance as imposed by conventional convolutional neural networks.
    We further accommodate different loss functions such as least squares for regression and cross-entropy minimization for classification using a loss-agnostic second order minimization framework.
\end{itemize}
Our approach leverages the exponentially higher theoretical expressivity of MPO over CPD, as shown in~\cite{oseledets2011tt}.
Furthermore, our MPO model generalizes the aforementioned $\Pi$‑nets without layer-wise non-linear activations and multivariate polynomial models by offering unifying architecture based on arbitrary rank decompositions and multilinear filters, such as convolution or (random) feature projections~\cite{karkarnick2012}.
Finally, compared to existing TT structures our approach is feature order independent where each block is associated to all the features as opposed to imposing feature specific blocks.
We evaluate the proposed (MPO)$^2$ structure for supervised learning on several tabular datasets and on image classification, and highlight its advantages over the latest tensor network based methods.

\section{Methods}
\subsection{Tensor Networks and Tensor Notation}
\begin{figure*}[h!]
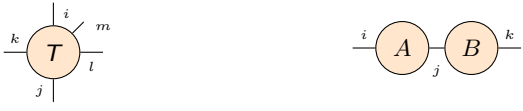
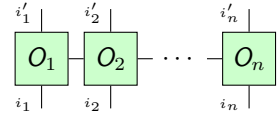

  \centering
  \subfloat[][5-d tensor $\protect\etT_{ijklm}$]{\parbox[b][1.9cm][c]{0.2\textwidth}{
      \centering
      \fivedtensor
  }}
  \hfill
  \subfloat[][Matrix product: $\protect(\mA \mB)_{ik} = \sum_j \protect\emA_{ij} \protect\emB_{jk}$]{\parbox[b][1.9cm][c]{0.33\textwidth}{
      \centering
      \matrixproduct
  }}
  \hfill
  \subfloat[][MPO: $\protect\etO^{\protect\vi^{n}\protect\vi'^{n}} = \sum_{\protect\va^{(n)}}
  O^{[1]i_1i_1'}_{1 a_2}O^{[2]i_2i_2'}_{a_2 a_3}\dots O^{[n]i_ni_n'}_{a_{n}1}$]{\parbox[b][1.9cm][c]{0.45\textwidth}{ \centering \protect\mpo }} \caption{ Graphical representation of (a) a tensor, (b) the product of two matrices and (c) the matrix product operator (MPO).
  }%
  \label{fig: tnexp}
\end{figure*}

Tensor networks are structures defined by a set of tensors and the dimensions of mutual contraction.
They are usually represented by graphs where each node is a tensor and the edges represent a contraction over a mode between the connected tensors.
Illustrative examples of tensor network based graphical representations can be seen in~\Figref{fig: tnexp}.
The figure represents, in the left panel, a tensor with five modes, in the middle panel a contraction of two tensors multiplied along one mode corresponding to conventional matrix multiplication and in the right panel the matrix product operator (MPO) corresponding to multiple tensors being pairwise contracted along one mode.

In this work, the position of the indices of a tensor, when referring to tensor networks structures, will be at the superscript to indicate vertical modes in the diagrammatic representation and at the subscript to indicate horizontal modes.
The different positions are mathematically equivalent, but conceptually, vertical modes are associated to the input space while horizontal modes are associated with the latent space.

Summation over multiple indices of the tensor $\tM$ with elements $\etM_{i_1i_2\dots i_n}$ will be denoted by $\sum_{i_1 i_2 \dots i_n} \etM_{i_1i_2\dots i_n}=\sum_{\vi^{(n)}}\etM_{\vi^{(n)}}$, where $\vi^{(n)} = \{\evi_1,\evi_2,\dots,\evi_n\}$, meaning that the sum is performed over all the indices going from $i_1$ to $i_n$.
When the superscript is omitted, it means that $\vi = \vi^{(N)}$, where $N$ is the degree of the polynomial.

The well-known tensor network structures, namely \textit{matrix product states}
(\textit{MPS}) or \textit{tensor trains} (\textit{TT}), as well as the Tucker and CPD
decompositions, are respectively given by

\begin{equation}
  \begin{aligned}
    \text{MPS/TT:} \, \etT_{\vd^{(N)}l}     & = \sum_{\vr}^{R}\etT^{[1]d_1}_{1
    r_2}\etT^{[2]d_2}_{r_2 r_3}\cdots \etT^{[n]d_n l}_{r_{n}1},                \\
    \text{Tucker/CPD:} \, \etT_{\vd^{(N)}l} & =
    \sum_{\vr}^{R}\mathcal{G}_{\vr l}\etT^{[1]d_1}_{r_1}\cdots
    \etT^{[n]d_n}_{r_{n}},
    \label{eq:MPSCPDTUCKER}
  \end{aligned}
\end{equation}

in which the Tucker decomposition reduces to the CPD when
$\mathcal{G}=\mathcal{I}$.\ \(\mathcal{I}\) is defined as the identity tensor with ones
along the (hyper-)diagonal and zeros elsewhere.
Notably, these decompositions are special cases of MPOs.
In~\Figref{fig:structures}~\subref{fig:structures-cpd} we provide a graphical representation of the Tucker/CPD structure contracted with a polynomial basis, while in~\subref{fig:structures-mps} we provide the graphical representation of MPS/TT contracted with a generic basis.
The generic basis representation for MPS/TT is the one most commonly found in the literature.

\subsection{Multivariate Polynomial Regression}
Given an input $\vx$ of dimension $D$, we define a multivariate polynomial of $\vx$ of
degree $N$:

\begin{equation}
  \begin{aligned}
    p_l(\vx) & = \etT_{l}^{(0)} + \sum_{d_1} \etT_{ld_1}^{(1)} \evx_{d_1} + \sum_{d_2 \geq d_1} \etT_{l\vd^{(2)}}^{(2)}\evx_{d_1}\evx_{d_2} + \\
    & + \cdots + \sum_{d_N \geq d_{N-1}}
    \etT_{l\vd^{(N)}}^{(N)} \evx_{d_1} \dots \evx_{d_N}
  \end{aligned}
  \label{eq: poly}
\end{equation}

where $l$ indicates one of the multivariate polynomial outputs, and $\etT$ are
the coefficients.

Following~\cite{Ayvaz-2022-CPD-StructuredMulti}, we consider two formulations of the polynomial.
Namely, as a sum of independent homogeneous polynomials of increasing
degree with coefficients parametrized as independent tensors for each degree
(type I), or with one tensor to represent all coefficients of the polynomial (type II):

\begin{align}
  \text{Type I:}  &  & p_l(\vx)         & = \sum_{n=0}^N \sum_{\vd^{(n)}}\etT_{l\vd^{(n)}}^{(n)}
  \evx_{d_1}\dots \evx_{d_n} \label{eq: degenaratepoly},                                         \\
  \text{Type II:} &  & p_l(\tilde{\vx}) & = \sum_{\vd^{(N)}}
  \tilde{\etT}_{l\vd^{(N)}}\tilde{\evx}_{d_1}\dots\tilde{\evx}_{d_N},
  \label{eq: homogeneouspoly}
\end{align}
where $\tilde{\bm{x}}=[1 ,\vx]$ is defined as the input vector $\vx$ with a constant additional
feature of value one (a bias term) that enables to account for all coefficients of all the different degrees of the polynomial.

Notably, the weight tensors grow exponentially in the number of coefficients as the degree $N$ of the polynomial increases for $M$ features by $\mathcal{O(M^N)}$ making the polynomial regression infeasible except at low degrees and with relatively few features.
To reduce the number of parameters the weight tensors have been decomposed using the CPD decomposition~\cite{hendrikx2019algebraic,Ayvaz-2022-CPD-StructuredMulti, govindarajan2022regression,chrysos2022augmenting, kilic2025interpretablebayesiantensornetwork} as well as tensor train decomposition~\cite{stoudenmire2016supervised,götte2021blocksparsetensortrainformat,kilic2025interpretablebayesiantensornetwork}.
However, the existing CPD procedures have limited modeling capacity whereas the existing TT modeling procedures~\cite{stoudenmire2016supervised,efthymiou2019tensornetworkmachinelearning,götte2021blocksparsetensortrainformat,kilic2025interpretablebayesiantensornetwork} are feature order dependent, decomposing the weight tensors using feature specific carts, i.e., $\etT^{[m]d'_{m-1}}_{r_{m-1} r_m}$, which is undesirable, as there often is no natural ordering of the features.
Such an ordering must therefore be engineered using domain knowledge or selected through heuristics, even though the quality of the TT representation may be sensitive to this choice.
As we will show, these drawbacks can be effectively addressed considering the MPO formalism.

\subsection{\({(\text{MPO})}^2\): Multivariate Polynomial optimization using
Matrix Product Operators}

Often in machine learning, to enhance the capability of the model, a linear transformation is applied to the inputs to learn suitable latent feature representations.
By applying a generic set of transformations $\mA^{[i]}$ to the
inputs, we can express the polynomial as:

\begin{equation}
  \begin{aligned}
    p_l(\vx) & = \sum_{\vd \vd'} \etT_{\vd' l}
    \emA^{[1]}_{d_1'd_1}\evx_{d_1}\dots\emA^{[N]}_{d_N'd_N}\evx_{d_N}
    \\
    & = \sum_{\vd \vd'} \etT_{\vd' l} \etA_{\vd'\vd}\evx_{d_1}\dots\evx_{d_N},
  \end{aligned}
  \label{eq: polylinear}
\end{equation}

\begin{figure*}[htb]
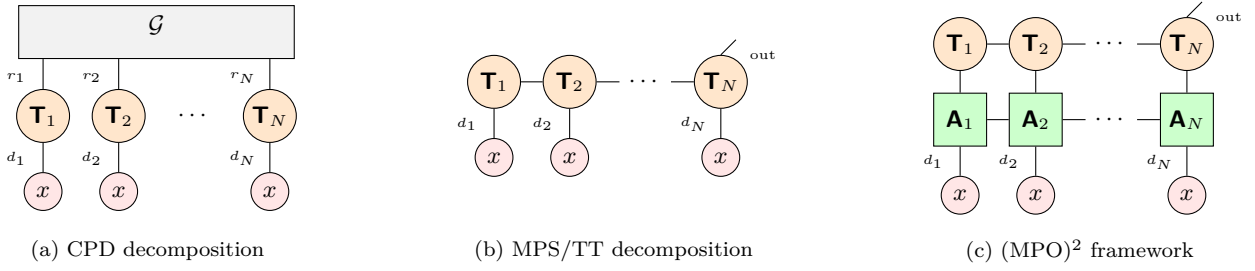

  \newcommand{\paraboxh}{3.2cm}

  \centering
  \subfloat[][CPD decomposition]{\parbox[b][\paraboxh][c]{0.32\textwidth}{
      \centering
      \cpd
  }\label{fig:structures-cpd}}
  \hfill
  \subfloat[][MPS/TT decomposition]{\parbox[b][\paraboxh][c]{0.32\textwidth}{
      \centering
      \mpstypeII
  }\label{fig:structures-mps}}
  \hfill
  \subfloat[][(MPO)$^2$ framework]{\parbox[b][\paraboxh][c]{0.32\textwidth}{
      \centering
      \mposqtypeII
  }\label{fig:structures-mpo}}

  \caption{
    Existing tensor network modeling procedures for multivariate polynomial
    regression based on (a) the CPD decomposition, (b) the MPS/TT
    decomposition and (c) the proposed (MPO)$^2$ framework exploring two layers of MPOs
    respectively transforming the input to suitable latent representations
    and creating a feature order invariant polynomial representation.
  }\label{fig:structures}
\end{figure*}
where we omit the superscript $\vd^{(N)}$ when it is equal to the degree of the
polynomial $N$.
The tensor product of all linear operators $\mA^{[i]}$ can be seen as a tensor $\tA$.

In the (MPO)$^2$ framework we propose to perform multivariate polynomial regression and classification by modeling both the generic linear transformation of the input space as well as the polynomial coefficients as \textit{matrix product operators} (MPOs).
These are diagrammatically represented in~\Figref{fig:structures}~\subref{fig:structures-mpo} and given as follows:

\begin{eqnarray}
  \etT_{\vd'^{(N)}l}  = \sum_{\vr}^{R}\etT^{[1]d'_1}_{1
  r_2}\etT^{[2]d'_2}_{r_2 r_3}\dots \etT^{[n]d'_n l}_{r_{n}1},\\
  \etA_{\vd'\vd} = \sum_{\va}^{R'}A^{[1]d_1d'_1}_{1 a_2}A^{[2]d_2d'_2}_{a_2
  a_3}\dots A^{[n]d_n d'_n}_{a_{n}1},
  \label{eq: MPO}
\end{eqnarray}
where $R$ and $R'$ are called respectively the rank of the coefficients and of the MPO structure.

\subsection{Three types of MPOs}
We propose three structures of the MPOs representing the input transformation tensor $\tA$ by linear projections, convolutions, and masking that accounts for weight redundancies by the proposed masking MPO.

In the following we outline the mentioned structures.

\subsubsection{Linear projection MPO (L-MPO)}
The following MPO represents an unstructured linear transformation of the input subspace, reducing the dimension from $D$ to $D'$, where $D' \ll D$.
As a result, we lower the complexity of inverting the Hessian during inference by a factor $\sim {\left(\frac{D'}{D}\right)}^{3}$.
The operator in its most general form as in~\Eqref{eq: polylinear} can be randomly initialized and learned blockwise in the same fashion as the structured polynomial coefficients are learned, in such a way that the model automatically infers the transformation of the inputs minimizing the loss.
Especially for high-dimensional inputs, computing and inverting the Hessian of a block can be challenging, and, if the inputs show linear dependency, wasteful.

We introduce a learnable operator that applies a simple linear transformation for each subspace represented by the blocks, which results in global structured linear transformation.
To further reduce the parameters, we can impose independency between the subspaces for the linear transformation by setting the rank of the MPO to one.
The advantage is that the new model, instead of representing the coefficients of the polynomial with blocks of dimension $R^2D$, instead is represented by two blocks of dimensions $R^2D'$ and $DD'$, where $D$ is the dimension of the input and $D'$ is the dimension of the projected subspace.
We define the linear MPO block as a randomly initialized learnable tensor $\etA^{d_i d_i'}_{a_i a_{i+1}}$.
When the dimension of the rank $a$ is $1$ we retrieve linear independent transformations of the inputs thereby transforming the weights of the model as

\begin{equation}
  \etT_{\vd}=\sum_{d_1'}(\etT^{(1)})_{d_1'}\emA_{d_1' d_1}\cdots \sum_{d_N'}(\etT^{(N)})_{d_N}\emA_{d_N' d_N}.
  \label{eq: weightlinear}
\end{equation}

\subsubsection{Convolution MPO (C-MPO)}
A structured case of linear projection are convolutions, which accounts for translation invariant compression as explored in CNNs~\cite{olshausen1996emergence}.
By representing the inputs as a two-dimensional tensor and projecting them along one of the two dimensions, we can derive an MPO acting as a convolution.

Specifically, if we consider images, we can define the two dimensions as patches and pixels in each patch, respectively called $p$ and $k$.
Consequently, the classical convolution can be written as

\begin{equation}
  \evx_p = \sum_k \evg_k \emX_{k p},
  \label{eq: appconvolvedinputs}
\end{equation}

and the resulting polynomial can be written as:

\begin{equation}
  \begin{aligned}
    & p(x_1, \dots, x_P) =
    \sum_{\vp}^P \sum_{\vr}^R\evx_{p_1}\etT^{p_1}_{1 r_2} \dots \evx_{p_n}\etT^{p_n}_{r_n 1} =                                                     \\
    & = \sum_{\vp}^P \sum_{\vr}\sum_{k_1}\evg_{k_1}\emX_{k_1 p_1}\etT^{p_1}_{1 r_2} \dots \sum_{k_n}\evg_{k_n}\emX_{k_n p_n}\etT^{p_n}_{r_n 1}\
  \end{aligned}
  \label{eq: convolvedpoly}
\end{equation}

Note that when the convolution kernels $\vg$ are different we cannot strictly speak of a polynomial with respect to the patches, since the inputs will be different in each block, but it will still be a polynomial over the full pixel space.

To rewrite~\Eqref{eq: convolvedpoly} as an MPO we can reorder the elements, add a summation over a delta function and vectorize the inputs to obtain

\begin{equation}
  \begin{aligned}
    & p(x_1, \dots, x_P) = \sum_{\vp}^P \sum_{\vp'}^P\sum_{\vr}\sum_{k_1}\emX_{k_1 p_1}\evg_{k_1}\delta_{p_1 p_1'}\etT^{p_1'}_{1 r_2} \cdots \\
    & \qquad\qquad \cdots \sum_{k_n}\emX_{k_n p_n}\evg_{k_n}\delta_{p_n p_n'}\etT^{p_n'}_{r_n 1}                                             \\
    & = \sum_\vk\sum_{\vp}^P \sum_{\vp'}^P\sum_{\vr}\evx_{(k_1,p_1)}\emA^{(k_1,p_1) p_1'}\etT^{p_1'}_{1 r_2} \cdots                                  \\
    & \qquad\qquad \cdots \evx_{(k_n,p_n)}\emA^{(k_n,p_n) p_n'}\etT^{p_n'}_{r_n 1},
  \end{aligned}
  \label{eq: convMPO}
\end{equation}

by defining the convolution block as:

\begin{equation}
  \etA^{(k_n,p_n) p_n'} = \evg_{k_n}\delta_{p_n p_n'}
  \label{eq: convMPOblock}
\end{equation}

where the index $(k,p)$ represents one index obtained by vectorizing over the
dimension in the parentheses, where $\delta$ is the delta function, taking
value one only if all indexes are the same and zero otherwise.
As a result, the MPO convolution block is defined as a linear projection on a subset of the full space.

Notably, using the MPO formalism it is natural to also increase the multiplicity of the kernels, by simply adding a bond dimension to the MPO block:

\begin{equation}
  \etA^{(k_n,p_n) p_n'}_{a_n a_{n+1}} = \sum_{a_n'}\etG^{k_n}_{a_n' a_{n+1}}\delta_{p_n p_n'}\delta_{a_n a_n'},
  \label{eq:convMPOblockrank}
\end{equation}
where \(\tG\) represents multiple (\(a_n \cdot a_{n+1}\)) kernels such that \(x_{p,a_n,a_{n+1}} = \sum_k\etG^{k}_{a_n a_{n+1}} \emX_{k,p}\), inducing interaction over different kernel subspaces.
Constructing the convolution MPO to accommodate higher dimensional inputs such as color channels in RGB images follows the same procedure.

\begin{figure*}[htb]
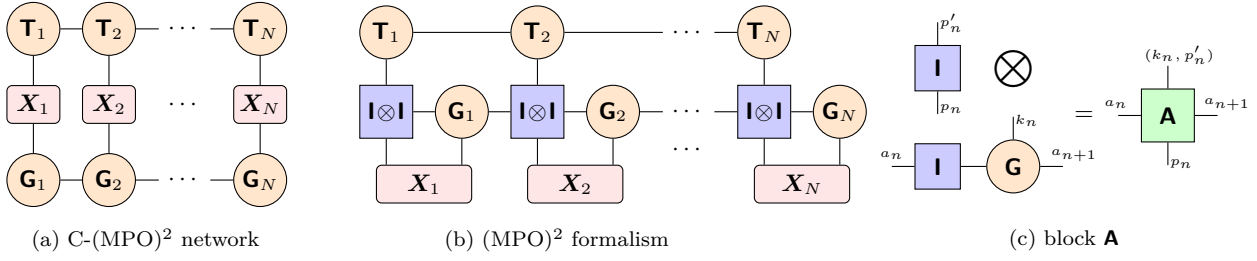

  \newcommand{\paraboxh}{3cm}

  \centering
  \subfloat[][C-(MPO)$^2$ network]{\parbox[b][\paraboxh][c]{0.26\textwidth}{
      \centering
      \cmpotwo
  }\label{fig:cmpotwo-net}}
  \hfill
  \subfloat[][(MPO)$^2$ formalism]{\parbox[b][\paraboxh][c]{0.29\textwidth}{
      \centering
      \cmpotwoasmpo
  }\label{fig:cmpotwo-mpo}}
  \hfill
  \subfloat[][block $\tA$]{\parbox[b][\paraboxh][c]{0.40\textwidth}{
      \centering
      \protect\cmpotwoasmpoblock
  }\label{fig:cmpotwo-block}}
  \caption{
    \protect\subref{fig:cmpotwo-net}
    The convolution (MPO)$^2$ network, with the top layer of coefficients $\tT$ acting on the convolved inputs $\mX_n$ through the kernels $\tG_n$.
    This form is the one used in the implementation of the code, since it requires contractions between smaller spaces.
    \protect\subref{fig:cmpotwo-mpo}
    The same network rewritten as an (MPO)$^2$: the middle layer is composed, for each column, of an identity block ($\tI\otimes\tI$) which connects horizontal legs of the MPO independently from the vertical ones.
    \protect\subref{fig:cmpotwo-block} convolution MPO block $\tA$
    of~\Eqref{eq:convMPOblockrank}.
  }\label{fig:cmpotwo}
\end{figure*}

The convolution can be seen as a linear projection on a subset of the full space.
We graphically represent in~\Figref{fig:cmpotwo} the convolutional (MPO)\(^2\) and how convolutions are included in the generic (MPO)\(^2\) block structure.

\subsubsection{Masking MPO (M-MPO)}
The existing tensor network based polynomial regression procedures have degenerate polynomial coefficients as defined in~\Eqref{eq: poly}, in which the weight tensor includes all orderings of multiplications of the same terms.

Notably, The number of coefficients for a multivariate polynomial scales as:

\begin{equation} N_{\text{Sym}} =
  \begin{pmatrix}
    M + D \\
    M
  \end{pmatrix}
  = \frac{(M+D)!}{D!
    M!
  }, \qquad N_{\text{Deg}} \approx M^D,
  \label{eq: sym}
\end{equation}

where the label Sym and Deg indicate Symmetric for the count when considering the symmetries between monomial and Degenerate when the symmetries are ignored.
Imposing symmetric constraint on the coefficients, is often hard to model, especially for tensor decomposition methods.
For this reason in the modeling of polynomial the symmetry is disregarded, leading to a number of represented parameters that in the worst case scales as depending on the model specifications.

Using Stirling's approximation for factorials we can calculate the fraction between $N_{\text{Deg}}$, the full degenerate space, and $N_{\text{Sym}}$, the non-degenerate space, defined in~\Eqref{eq: sym} as:

\begin{equation}
  \frac{N_{\text{Deg}}}{N_{\text{Sym}}} := K \simeq \left(1+\frac{n}{d}\right)^{-(d+n+\frac{1}{2})}n^{n +\frac{1}{2}}\sqrt{2\pi},
  \label{eq: fractionbehaviour}
\end{equation}

with $b,n \gg 1$ (for $d>8$ the approximation is already valid).
Fixing $n$ the limit behavior of $K$ respect to $d$ is $\lim_{d \rightarrow \infty} K \simeq e^{-n}n^n$ (which for $n=6$ is $\simeq 115$ and for $n=10$ it is $\sim \nicefrac{1}{2}10^5$).
Consequently, for large polynomial degrees $n$, the divergence of $K$ can impair the expression power of the model and definitely hinder the explainability of the model.

The ideal scenario would be to associate each input combination (monomial) to one and only one element of the learned coefficients tensor.
We can achieve this by introducing a mask that allows non zero connection between input and model for only one monomial for each set of equivalent monomials, obtaining the natural polynomial definition:

\label{subsec:cumsumop}
\begin{equation}
  p = \sum_{\vr}^{r}\sum_{d_1\geq 0}^D \etT^{d_1}_{1  r_2}\tilde{x}_{d_1}\sum_{d_2\geq d_1}^D \etT^{d_2}_{r_2  r_3}\tilde{x}_{d_2}\dots\sum_{d_n\geq d_{n-1}}^D \etT^{d_n}_{r_n  1}\tilde{x}_{d_n}
  \label{eq: polynomialcumsum}
\end{equation}

\begin{figure*}[t]
  \newcommand{\paraboxh}{2.4cm}

  \centering
  \subfloat[][masking MPO]{\parbox[b][\paraboxh][c]{0.45\textwidth}{
      \centering
      \cumsummpo
  }\label{fig:cummpoblock-op}}
  \hfill
  \subfloat[][masking MPO block]{\parbox[b][\paraboxh][c]{0.45\textwidth}{
      \centering
      \cumsumblock
  }\label{fig:cummpoblock-block}}
  \caption{
    \protect\subref{fig:cummpoblock-op}
    Diagrammatic representation of the masking MPO in~\Eqref{eq: polynomialoperatorcumsum}.
    \protect\subref{fig:cummpoblock-block}
    Diagrammatic representation of a block of the masking MPO block in~\Eqref{eq: cumsummpo}.
  }
  \label{fig: cummpoblock}
\end{figure*}

We wish to rewrite the masking action in the form of an MPO thereby enalbing to retain the block structure of the problem needed to use the block-wise learning algorithm.
We define two auxiliary tensors, the Heaviside matrix \(\bm{\Theta}\), and the hyperdiagonal tensor \(\mathcal{I}\).

\begin{equation}
  \begin{aligned}
    & \bm{\Theta}_{i j} = \theta(j-i)              \\
    & \mathcal{I}_{a b \dots} = \delta_{a b \dots}
  \end{aligned}
  \label{eq: heavisidehyperdiag}
\end{equation}

Where $\theta$ represents the Heaviside function, where \(\theta(x) = 1 \quad if \quad x \geq 0 \quad \theta(x) = 0\) otherwise and $\delta$ is a function that is $1$ only if all indices are the same and $0$ otherwise.

\Eqref{eq: polynomialcumsum} can be rewritten:

\begin{equation}
  \begin{aligned}
    & p(x_1,\dots,x_n) = \sum_{\vr}^{r}\sum_{\va}^{D}\sum_{d_1, k, d_1'}^D \etT^{d_1}_{1  r_2} \bm{\Theta}_{0  k}\mathcal{I}_{k d_1 d_1'a_2}\tilde{x}_{d_1'}\cdots \\
    & \qquad \cdots\sum_{d_n,k,d_n'}^D \etT^{d_n}_{r_n  1}\bm{\Theta}_{a_n k}\mathcal{I}_{k  d_n d_n'}\tilde{x}_{d_n'}.
  \end{aligned}
  \label{eq: polynomialoperatorcumsum}
\end{equation}

We can now extract an MPO by defining its blocks.

\begin{equation}
  \etA^{d_i d_i'}_{a_i a_i+1} = \sum_k \bm{\Theta}_{a_i k}\mathcal{I}_{k d_i  d_i' a_{i+1}},
  \label{eq: appcumsumblocks}
\end{equation}
diagrammatically represented in~\Figref{fig: cummpoblock}. Finally, we obtain the masking MPO as

\begin{equation}
  \etA_{\vd'}^{\vd} = \sum_{a_2, \dots, a_n}\etA^{d_1 d_1'}_{1 a_2}\etA^{d_2 d_2'}_{1 a_2}\dots \etA^{d_n d_n'}_{a_n 1},
  \label{eq: cumsummpo}
\end{equation}
represented in~\Figref{fig: cummpoblock}.

The polynomial can now be expressed as a contraction between a tensor representing coefficients, a mask given by the masking MPO and a tensor containing the inputs, all retaining the block structure.

\begin{equation}
  p(x_1,\dots,x_n) = \sum_{\vd'}^D\sum_{\vd}^D \etT^{\vd'}\etA^{\vd}_{\vd'}X_{\vd}
  \label{eq: polympocumsum}
\end{equation}

Notably, due to the separation of the masking MPO, the gradient and subsequent Hessian calculations remain unchanged.

\subsection{A fourth structure, the Ring}

A direct consequence of the permutation invariance of the polynomial input map used in~\Eqref{eq: homogeneouspoly}, and the linearity of the model is that the coefficients of the optimal solution are permutation invariant with respect to a basis change, belonging to the fully symmetric space.

In~\cite{ayvaz2021eusipco} symmetric CPD structures are explored, where symmetry (and permutation invariance) is obtained by repeating an identical block through the input space.
On the same note, we define the respective counterpart for MPS/MPO formalism, the invariant ring.

Due to the natural structural asymmetry of MPS structures, to obtain permutation invariance imposing all blocks identical is not enough.
We need to impose also periodic boundary conditions,
obtaining the invariant ring defined as:

\begin{equation}
  \etR^{i_1 \ldots i_N} = \sum_{r_1 \dots r_{N}}
  \etT_{r_1 r_2}^{i_1} \cdots \etT_{r_N r_1}^{i_N} \in \mathbb{R}^{\times_{j=1}^N \dim(i_j)}.
  \label{eq: ring}
\end{equation}

The contraction with the input space results in an elegant and simple formulation:

\begin{equation}
  \sum_{\bm{i}}\etR^{i_1 \ldots i_N} \evx_{i_1} \ldots \evx_{i_N} = trace \left[{\left( \sum_i \mT^i \evx_i\right)}^N\right].
  \label{eq: forwardring}
\end{equation}

Note that the ring as defined is an element of the translation invariant space, meaning invariant to cyclic permutations of the axis.
The fully symmetric space is a smaller subspace residing entirely inside the cyclic space. We could impose even more constraint to ensure that the ring belongs to the fully symmetric space, invariant to all permutations of the axis, which is that all rank by rank matrices for each input dimension commute with each other.

Utilizing the ring structure enables to take advantage of the permutation invariant property of the solution to further reduce the parameters of the model without compromising expressivity.

The invariant ring structure is not trivially learnable using natural gradient, because the derivative with respect to a block belongs to an asymmetric, non-linear space, causing the second derivative to be non-trivial to compute.
As a result, we only use gradient based procedures to learn the parameters of the ring.

\subsection{Alternating natural gradient}\label{sec: natural}

Natural gradient is a second-order optimization method that calculates the update step of the parameters taking into consideration the curvature of the loss, resulting in faster convergence with respect to the number of steps.
Often utilized algorithms for MPOs are alternating least squares (ALS)~\cite{ALS} or the density matrix renormalization group (DMRG)~\cite{SCHOLLWOCK201196}, both sharing similar computational properties and methods.

Given an objective loss to minimize $\min_{\theta}L(y, x(\theta))$, natural gradient defines the best update of the parameters as $ \Delta \theta = -\mH^{-1}_{\theta}(L)\vj_{\theta}(L) $, where $\mH_{\theta}(L)$ and $\vj_{\theta}(L)$ are the Hessian and the Jacobian of the loss with respect to the parameters.
Inspired by alternating linear systems, also denoted alternating least squares, (ALS) methodologies on tensor networks, we learn the update step block-wise.
The method reduces to computing the Hessian and Jacobian of the loss with respect to a block, update the block according to the step, and repeat the process until all blocks are updated and then proceed to repeat the operation in the opposite direction.
We denote the full iteration as a \emph{sweep}.

The Hessian is often singular in the first sweep due to random initialization, especially when considering losses other than least squares minimization.
To stabilize the inference, we apply Tikhonov regularization~\cite{boyd2009exponentially, trefethen2019approximation}, with an exponentially decaying schedule for weight decay.

Importantly, for MPO structures, calculating the Hessian of a single block reduces to a trivial task.
We can write the Hessian and Jacobian taking into account the
regularization:

\begin{equation*}
  \begin{aligned}
    & \vj_{\tA^{(i)}}(L) = \sum_{sl} \nabla_{\tA^{(i)}} p_{ls}
    \;\partial_{p_{ls}}L(\vp_s,y_s) + \lambda \tA^{(i)},
    \\
    & \mH_{\tA^{(i)}}(L) =
    \sum_{sll'}\nabla_{\tA^{(i)}}p_{ls}\;\nabla^T_{\tA^{(i)}}p_{l's} \;
    \partial_{p_{ls}} \partial_{p_{l's}} L(\vp_s,\vy) + \lambda \mI,
  \end{aligned}
  \label{eq:HessianJacobianTrain2}
\end{equation*}
where $p_{ls}$ is the output of the model for sample $s$ and output dimension
$l$ and $\vp_s$ is the vector of outputs for sample $s$.

Block-wise learning and MPO structured coefficients simplify the Hessian since \mbox{$\nabla_{\tA^{(i)}} \nabla_{\tA^{(i)}} p=0$}.
Additionally, the gradient with respect to a block amounts to calculating the contraction of the full MPO without the differentiated block.
For computation we replace the \(\lambda\) regularization scale, whit an effective \(\lambda_e\), obtained by multiplying to \(\lambda\) the trace of the absolute Hessian without regularization.

Notably when using the least squares loss the natural gradient method is equivalent to the more commonly used ALS method defined in~\cite{Holtz-2012-TheAlternatingLine}.

\subsection{Complexity analysis}\label{sec:complexity}
In order to do a complexity analysis of the different models, we distinguish the two main contributions of complexity, the contraction of the network and the Newton optimization step.
As for the contraction cost of the network we will use a generic optimal path for contraction, which in practice can change due to different relative dimensions.
While the cost of the construction and inversion of Hessian depends exclusively on the dimensions of the node.

The Hessian is constructed by multiplying over a sample dimension the outer product of the environment of a block.
Let us consider a generic block that can describe all blocks needed for models presented in the paper, \(\tA \in \sR^{u \times d \times l \times r \times o}\), where for consistency with graphical notation we can think at the modes respectively as up, down, left, right and out.
naming \(U = \dim{(u)}\dim{(d)}\dim{(l)}\dim{(r)}\), the complexity \(C_H\) of computing the Hessian for \(\tA\) is:

\[
  C_H = U \cdot S \cdot U,
\]
where \(S\) is the sample dimension.

The complexity \(I_H\) of solving the linear system for \(\tA\) considering that the Hessian is symmetric and we use Cholesky decomposition is:

\[ I_H = \frac{1}{3} U^3, \]

and each block will contribute to the complexity with a factor \(\simeq \frac{1}{3}U^3 + S\cdot U^2\).

As for the contraction strategy, we can view our type of models as networks with columns and rows, as it can be identified in~\Figref{fig:structures}~\subref{fig:structures-mpo}.
Commonly we start from the left and contract all vertical blocks in the first column.
Then we contract the next column adding one row at a time.
We consider to contract one sample at a time, and performing as initial contraction the one between the input and its associated block.
With the previously constructed contraction, we can see that the complexity of contracting a block, is maximized by \(C_F = U \cdot r\), where \(r\) is the biggest horizontal dimension.
Then for each sample the full complexity cost is \(C_F = U \cdot r \cdot S\).
For CPD the cost differs, since the inputs are contracted with the respective block, and then elementwise multiplied, leading to a cost linear in the rank and the number of blocks and it will be \(C_F^{CPD} = C_F/r\).
The full maximal cost calculation that accounts for forward contractions and node update calculation will be approximated as

\[ C = \frac{1}{3}U^3 + S \cdot U^2 + U \cdot r \cdot S.
\]

We evaluate the complexity \(C\) for all models, and perform the calculation with respect to the block with higher dimensionality.

In the following, we indicate with \(d\) the number of features of the dataset and by \(N\) the degree of the polynomial modeled.
The actual input dimension will often be \(\hat{d} = d + 1\), which accounts for the added bias term to the feature vector.

\subsubsection{MPS}

For the standard model with no MPO between the polynomial coefficients and the data, we can substitute \(U = r^2 \hat{d}\), and observe that we will have number of blocks equal to the degree of the polynomial \(N\)

\[ \frac{C_{MPS}}{N} = \frac{1}{3}r^6\hat{d}^3 + S r^4 \hat{d}^2 + S r^3 \hat{d}.
\]

Notably, contrary to our MPS formulation the tensor network for machine learning (TNML) models proposed in \cite{efthymiou2019tensornetworkmachinelearning, götte2021blocksparsetensortrainformat} has the number of blocks correspond to the number of features, while the dimension of the input mode, is two for the Fourier basis (TNML-F)~\cite{efthymiou2019tensornetworkmachinelearning} and degree plus one for the polynomial basis (TNML-P) ~\cite{götte2021blocksparsetensortrainformat} Consequently, they have the same complexity form as MPS:

\[ C_{TNML-F} = d\left(\frac{1}{3}r^6 2^3 + S r^4 2^2 + S r^3 2\right), \]

\[ C_{TNML-P} = d\left(\frac{1}{3}r^6 \hat{N}^3 + S r^4 \hat{N}^2 + S r^3 \hat{N}\right), \]

where \(\hat{N} = N + 1\).

\subsubsection{L-(MPO)\(^2\)}

As for standard linear projection, we consider that the MPO is simply reducing the dimension of each input from \(\hat{d}\) to \(d'\), and have rank \(1\).
We will then have \(N\) blocks of dimension \(d'\) and \(N\) blocks of dimension \(\hat{d}d'\), obtaining

\[ \frac{C_L}{N}= \frac{1}{3}d'^3 \left(r^6 + \hat{d}^3\right) + S d'^2 \left(r^4 + \hat{d}^2\right) + S d' r \left( r^2 + \hat{d}\right).
\]

\subsubsection{C-(MPO)\(^2\)}

Depending on the input shape, the convolutional model will contribute to the complexity as one MPS for each mode as can be seen from~\Figref{fig:cmpotwo}.
As a result, parameterizing the cost \(C_{MPS} (\hat{d}_m, r_m)\) in function of the rank and dimension of the MPO associated to each mode of the inputs.
For an input structured with \(M\) modes:

\[ C_C = \sum_{m=0}^M C_{MPS} (\hat{d}_m, r_m).
\]

\subsubsection{M-(MPO)\(^2\)}

The Masking operator is fixed, and so will bring a contribution to the complexity only in the contraction obtaining.
Still, it can amount to a large increase in the cost for high dimensional data.

\[
  \frac{C_M}{N}= \frac{1}{3}\hat{d}^3 r^6 + S \hat{d}^2r^4 + S r \hat{d}\left(r^2 + \hat{d}^4\right).
\]

\subsubsection{CPD}

The CPD dimension for a block is linear instead of quadratic with respect to a rank, and the contraction of the model is done through element-wise multiplications, obtaining a complexity of

\[ \frac{C_{CPD}}{N} = \frac{1}{3}r^3\hat{d}^3 + S r^2 \hat{d}^2 + S r \hat{d}.
\]

\section{Experimental Setup}\label{sec:experiments}
We compare our propsed (MPO)$^2$ to CPD for polynomial regression using symmetric CPD based on TeMPO~\cite{Ayvaz-2022-CPD-StructuredMulti} and asymmetric CPD (CPD-A)~\cite{govindarajan2022regression} optimized in our framework.
We further include the classical TT/MPS structure for regression both with Fourier basis (TNML-F) as in~\cite{efthymiou2019tensornetworkmachinelearning} and polynomial basis (TNML-P) as in~\cite{götte2021blocksparsetensortrainformat}.
For comparison, we also included Gaussian Processes (GP) and XGBoost~\cite{chen2016xgboost} as implemented in scikit-learn~\cite{scikit-learn} as well as a multilayer perceptron (MLP) and the Base mean estimator model predicting based on the training set average outputs.

The datasets are chosen based on popularity in the UCML repository~\cite{kelly2019uciml}.
The historical views can be seen through these links:\footnote{Classification: \url{https://web.archive.org/web/20250923141238/https://archive.ics.uci.edu/datasets?
Task=Classification}}\textsuperscript{,}\footnote{Regression: \url{https://web.archive.org/web/20250923141141/https://archive.ics.uci.edu/datasets?
Task=Regression}}

The data and pre-processing pipeline is as follows for all datasets, and all datasets were processed using the same method:

Feature columns containing missing values are removed.
Targets are treated as a single column and converted to integer labels for classification tasks.

Feature encoding follows a capped one hot scheme to control dimensionality.
Numeric columns are always kept.
Categorical columns are one hot encoded, subject to a fixed maximum number of total feature columns.
Columns with the largest cardinality are dropped first if the budget is exceeded.
If the number of one hot encoded columns still exceeds the budget, excess variables are trimmed.

Data is split into training, validation, and test sets with proportions of 70\%, 15\%, and 15\%, respectively.
Standard normalization is used by finding the mean and standard deviation on the training set's numeric columns and applied to the corresponding validation and test columns, while one hot features remain unchanged.

Some datasets were discarded based on not fitting into this general data pipeline.
We discarded the dataset diabetes\footnote{\url{https://archive.ics.uci.edu/dataset/34/diabetes}} due to unavailability for download through the official package.
We discarded Automobile\footnote{\url{https://archive.ics.uci.edu/dataset/10/automobile}} and Auto MPG\footnote{\url{https://archive.ics.uci.edu/dataset/9/auto+mpg}} due to the small number of instances together with the presence of missing data.

Details of the number of samples and features are provided in table~\ref{tab: dataset detail}.
\begin{table}[!t]
  \tiny
  \caption{
    Datasets with tasks, sizes, features, and shorthand codes.
  }\label{tab: dataset detail}
  \begin{tabular}{ccllrrrr}
    \toprule
    Code & Dataset                                                                       & Task & Train & Val/Test & Feat. \\
    \midrule
    AD   & \href{https://archive.ics.uci.edu/dataset/2}{Adult}                           & C    & 34189 & 7326     & 47    \\
    BA   & \href{https://archive.ics.uci.edu/dataset/222}{Bank Marketing}                & C    & 31647 & 6782     & 33    \\
    MU   & \href{https://archive.ics.uci.edu/dataset/73}{Mushrooms}                      & C    & 5686  & 1219     & 50    \\
    WQ   & \href{https://archive.ics.uci.edu/dataset/186}{Wine Quality}                  & C    & 4547  & 975      & 11    \\
    SD   & \href{https://archive.ics.uci.edu/dataset/697}{Students' Dropout}             & C    & 3096  & 664      & 36    \\
    CE   & \href{https://archive.ics.uci.edu/dataset/19}{Car Evaluation}                 & C    & 1209  & 259      & 27    \\
    BR   & \href{https://archive.ics.uci.edu/dataset/17}{Breast Cancer Wisconsin}        & C    & 398   & 85       & 30    \\
    HE   & \href{https://archive.ics.uci.edu/dataset/45}{Heart Disease}                  & C    & 212   & 45       & 11    \\
    WI   & \href{https://archive.ics.uci.edu/dataset/109}{Wine}                          & C    & 124   & 27       & 13    \\
    IR   & \href{https://archive.ics.uci.edu/dataset/53}{Iris}                           & C    & 105   & 22       & 4     \\
    PO   & \href{https://archive.ics.uci.edu/dataset/332}{Online News Popularity}        & R    & 27750 & 5947     & 58    \\
    AP   & \href{https://archive.ics.uci.edu/dataset/374}{Appliances Energy Prediction}  & R    & 13814 & 2961     & 27    \\
    BK   & \href{https://archive.ics.uci.edu/dataset/275}{Bike Sharing}                  & R    & 12165 & 2607     & 12    \\
    AI   & \href{https://archive.ics.uci.edu/dataset/601}{AI4I}                          & R    & 7000  & 1500     & 9     \\
    SB   & \href{https://archive.ics.uci.edu/dataset/560}{Seoul Bike Sharing}            & R    & 6132  & 1314     & 18    \\
    AB   & \href{https://archive.ics.uci.edu/dataset/1}{Abalone}                         & R    & 2923  & 627      & 11    \\
    OB   & \href{https://archive.ics.uci.edu/dataset/544}{Obesity Levels}                & R    & 1477  & 317      & 39    \\
    CO   & \href{https://archive.ics.uci.edu/dataset/165}{Concrete Compressive Strength} & R    & 721   & 155      & 8     \\
    EE   & \href{https://archive.ics.uci.edu/dataset/242}{Energy Efficiency}             & R    & 537   & 116      & 8     \\
    SP   & \href{https://archive.ics.uci.edu/dataset/320}{Student Performance}           & R    & 454   & 98       & 50    \\
    RE   & \href{https://archive.ics.uci.edu/dataset/477}{Real Estate Valuation}         & R    & 289   & 63       & 6     \\
    \bottomrule
  \end{tabular}
\end{table}

\subsection{Hyperparameter search}\label{sec:hyperparameters}
We conduct a hyperparameter grid search on the validation set for all models reported and describe below the range and model types for these searches.

\subsubsection{(MPO)$^2$}
Due to the Hessian being unstable in the early phase of optimization, we applied Tikhonov regularization with an exponentially decaying schedule.
To find the suitable regularization level we decay it and use early stopping to stop when validation loss does not decrease for ten block/operator updates.
In all tabular experiments we start with an initial value of $\lambda_{\text{start}}=5$ and decay exponentially with $\gamma=0.25$ as such: $\lambda_n=\lambda_{\text{start}}\cdot \gamma^n=5.0\cdot{(0.25)}^n$ where $n$ is the number of sweeps done.

Apart from training using the natural gradient procedure, we additionally considered training using gradient descent with weight decay, AdamW~\cite{DBLP:journals/corr/abs-1711-05101}.
The optimizer hyperparameters were fixed across all runs: a learning rate of
\(0.005\), the AdamW optimizer, and a weight decay of \(0.01\). Training ran for a maximum of
\(1000 \) epochs with batch size of \(512\) and early stopping with a patience of \(100\)
epochs and a minimum improvement threshold of \(0.001\).  The model learned through the natural gradient method described in Section~\ref{sec: natural} will be indicated with N, while the one learned with gradient descent will be indicated by G.

For stability, we multiply the regularization constant by the mean of the absolute values in the Hessian diagonal.

The TNML-F models are optimized using the present optimization framework to directly assess the impact on model structure on performance as well as our implementation of the original paper density matrix renormalization group (DMRG) based gradient method.

For TNML models, we ablate over a set hyperparameters as well as different methods and report the test result for the best performing configuration.
TNML-P is learned both with our implementation of ALS, as in the original paper, and gradient descent.
Additionally, for TNML-F we implemented the training method described in~\cite{stoudenmire2016}, which uses DMRG-inspired gradient descent.

\begin{table*}[ht]
\centering
\small
\caption{Regression results.}
\label{tab:regression_results}
\begin{tabular}{lcccccccccc}
\toprule
& RE & EE & CO & SP & OB & AB & SB & AI & BK & PO \\
\midrule
\textbf{N-(MPO)}$\bm{^2}$ & 78.44 & \underline{99.77} & 84.03 & 21.81 & \underline{73.03} & 59.87 & 66.88 & \underline{41.39} & 67.08 & 2.34 \\
 & $\pm$3.07 & $\pm$0.00 & $\pm$1.57 & $\pm$0.54 & $\pm$1.07 & $\pm$0.19 & $\pm$2.16 & $\pm$0.30 & $\pm$0.09 & $\pm$0.52 \\
\midrule
\textbf{G-(MPO)}$\bm{^2}$ & 81.22 & 99.58 & 84.29 & \textbf{23.07} & 68.74 & \underline{60.16} & \underline{73.65} & 40.75 & 66.81 & \textbf{3.75} \\
 & $\pm$1.14 & $\pm$0.04 & $\pm$1.56 & $\pm$1.74 & $\pm$2.72 & $\pm$0.42 & $\pm$0.44 & $\pm$1.89 & $\pm$0.17 & $\pm$0.42 \\
\midrule
\textbf{G-Ring} & \underline{81.49} & 99.48 & 84.19 & 20.04 & 66.57 & 59.48 & 72.10 & 38.85 & 66.41 & 1.45 \\
 & $\pm$3.39 & $\pm$0.07 & $\pm$1.25 & $\pm$7.11 & $\pm$2.77 & $\pm$0.29 & $\pm$1.82 & $\pm$2.98 & $\pm$0.08 & $\pm$1.10 \\
\midrule
\midrule
\textbf{N-CPD-A} & 80.61 & 99.52 & 82.37 & 20.30 & 71.35 & 59.67 & 60.75 & 39.07 & 66.33 & 1.76 \\
 & $\pm$0.78 & $\pm$0.17 & $\pm$1.78 & $\pm$2.02 & $\pm$1.28 & $\pm$0.68 & $\pm$6.14 & $\pm$1.79 & $\pm$0.15 & $\pm$0.99 \\
\midrule
\textbf{G-CPD-A} & 77.90 & 99.50 & \underline{85.05} & 21.06 & 64.12 & 59.68 & 62.37 & 39.03 & 66.51 & 2.51 \\
 & $\pm$11.40 & $\pm$0.12 & $\pm$1.12 & $\pm$2.34 & $\pm$1.97 & $\pm$0.56 & $\pm$8.22 & $\pm$1.80 & $\pm$0.15 & $\pm$0.50 \\
\midrule
\textbf{TEMPO} & 10.25 & 92.65 & 49.36 & 19.60 & 51.86 & 56.19 & 44.26 & 31.06 & 41.45 & 1.13 \\
 & $\pm$16.41 & $\pm$0.38 & $\pm$18.37 & $\pm$7.49 & $\pm$1.55 & $\pm$2.37 & $\pm$17.01 & $\pm$5.31 & $\pm$0.38 & $\pm$0.54 \\
\midrule
\midrule
\textbf{TNML-P} & F & 99.65 & 71.46 & F & F & F & 71.85 & 28.78 & \underline{70.58} & F \\
 &  & $\pm$0.13 & $\pm$7.48 &  &  &  & $\pm$0.46 & $\pm$1.83 & $\pm$1.77 &  \\
\midrule
\textbf{TNML-F} & F & F & F & F & F & F & -97.52 & -1.34 & F & -8.99 \\
 &  &  &  &  &  &  & $\pm$0.97 & $\pm$0.28 &  & $\pm$0.03 \\
\midrule
\midrule
\textbf{MLP} & 27.24 & 92.22 & 64.13 & 18.42 & 89.33 & 58.00 & 92.24 & \textbf{60.58} & 94.25 & 2.34 \\
 & $\pm$1.91 & $\pm$0.19 & $\pm$0.69 & $\pm$0.74 & $\pm$2.68 & $\pm$1.57 & $\pm$1.30 & $\pm$2.14 & $\pm$0.45 & $\pm$0.03 \\
\midrule
\textbf{XGBoost} & \textbf{82.48} & \textbf{99.83} & \textbf{92.06} & 19.61 & 92.12 & 55.41 & \textbf{97.81} & 59.12 & \textbf{94.84} & 0.72 \\
\midrule
\textbf{GP} & 70.99 & 99.79 & 85.88 & 21.97 & \textbf{94.22} & \textbf{60.40} & -- & -- & -- & -- \\
\midrule
\textbf{Base} & -0.46 & -0.09 & -0.00 & -6.30 & -0.59 & -0.08 & -0.01 & -0.15 & -0.06 & -0.03 \\
\bottomrule
\end{tabular}
\end{table*}

\begin{table*}[ht]
\centering
\small
\caption{Classification results.}
\label{tab:classification_results}
\begin{tabular}{lcccccccccc}
\toprule
& IR & HE & WQ & BR & AD & BA & WI & CE & SD & MU \\
\midrule
\textbf{N-(MPO)}$\bm{^2}$ & 97.83 & 61.52 & 55.74 & \textbf{99.42} & \underline{57.08} & 84.33 & 99.26 & \textbf{99.50} & 76.43 & \textbf{99.51} \\
 & $\pm$3.07 & $\pm$3.08 & $\pm$0.60 & $\pm$0.82 & $\pm$0.07 & $\pm$7.69 & $\pm$1.56 & $\pm$0.70 & $\pm$0.75 & $\pm$0.00 \\
\midrule
\textbf{G-(MPO)}$\bm{^2}$ & 99.13 & \textbf{65.65} & \underline{55.95} & 98.84 & 56.89 & \underline{90.54} & \textbf{100.00} & 99.38 & 76.55 & 99.10 \\
 & $\pm$1.83 & $\pm$4.20 & $\pm$0.86 & $\pm$0.95 & $\pm$0.13 & $\pm$0.13 & $\pm$0.00 & $\pm$0.37 & $\pm$0.74 & $\pm$0.28 \\
\midrule
\textbf{G-Ring} & 99.57 & 65.43 & 55.09 & 98.72 & 56.58 & 90.43 & 95.19 & 97.38 & 75.05 & 99.10 \\
 & $\pm$1.37 & $\pm$3.47 & $\pm$1.10 & $\pm$1.16 & $\pm$0.17 & $\pm$0.16 & $\pm$2.50 & $\pm$0.35 & $\pm$0.67 & $\pm$0.28 \\
\midrule
\midrule
\textbf{N-CPD-A} & 99.57 & 60.65 & 54.69 & 99.30 & 56.88 & 75.75 & 99.26 & 98.15 & \underline{77.55} & \textbf{99.51} \\
 & $\pm$1.37 & $\pm$2.98 & $\pm$2.11 & $\pm$0.98 & $\pm$0.07 & $\pm$15.86 & $\pm$1.56 & $\pm$0.40 & $\pm$0.54 & $\pm$0.00 \\
\midrule
\textbf{G-CPD-A} & 99.13 & 60.00 & 55.55 & 99.07 & 56.57 & 90.49 & \textbf{100.00} & 98.12 & 75.81 & 99.06 \\
 & $\pm$1.83 & $\pm$2.93 & $\pm$1.00 & $\pm$0.92 & $\pm$0.20 & $\pm$0.18 & $\pm$0.00 & $\pm$0.34 & $\pm$0.91 & $\pm$0.27 \\
\midrule
\textbf{TEMPO} & \textbf{100.00} & 58.70 & 54.13 & 96.51 & 56.57 & 90.16 & \textbf{100.00} & 82.15 & 76.87 & 99.13 \\
 & $\pm$0.00 & $\pm$2.17 & $\pm$1.39 & $\pm$3.08 & $\pm$0.31 & $\pm$0.28 & $\pm$0.00 & $\pm$3.12 & $\pm$0.86 & $\pm$0.10 \\
\midrule
\midrule
\textbf{TNML-P} & \textbf{100.00} & 14.57 & 16.62 & -- & 25.43 & 50.56 & 75.19 & 20.92 & 33.37 & 47.33 \\
 & $\pm$0.00 & $\pm$11.32 & $\pm$5.03 &  & $\pm$5.00 & $\pm$1.66 & $\pm$12.23 & $\pm$6.24 & $\pm$2.92 & $\pm$11.64 \\
\midrule
\textbf{TNML-F} & 60.87 & 32.61 & 34.11 & 50.47 & 33.18 & 49.81 & 35.19 & 24.35 & 34.49 & 49.29 \\
 & $\pm$0.00 & $\pm$5.80 & $\pm$3.11 & $\pm$6.42 & $\pm$1.50 & $\pm$0.57 & $\pm$11.61 & $\pm$7.64 & $\pm$3.46 & $\pm$9.27 \\
\midrule
\midrule
\textbf{MLP} & 97.39 & 53.48 & 60.00 & 99.30 & 57.10 & \textbf{90.96} & 99.26 & 98.46 & 76.51 & 99.46 \\
 & $\pm$3.89 & $\pm$1.19 & $\pm$2.01 & $\pm$1.04 & $\pm$0.13 & $\pm$0.07 & $\pm$1.66 & $\pm$0.77 & $\pm$0.51 & $\pm$0.04 \\
\midrule
\textbf{XGBoost} & \textbf{100.00} & 54.35 & \textbf{67.79} & 96.51 & \textbf{57.81} & 90.95 & \textbf{100.00} & 96.54 & 78.01 & \textbf{99.51} \\
\midrule
\textbf{GP} & \textbf{100.00} & 58.70 & 61.03 & 97.67 & -- & -- & 96.30 & 96.54 & \textbf{78.31} & -- \\
\midrule
\textbf{Base} & 43.48 & 54.35 & 46.46 & 69.77 & 50.91 & 88.54 & 37.04 & 69.23 & 50.00 & 50.94 \\
\bottomrule
\end{tabular}
\end{table*}

\subsubsection{Gaussian Process (GP)}
We evaluated 14 Gaussian process kernel configurations for datasets with fewer than 4000 samples.
The base kernels included a radial basis function (RBF) kernel, a Matérn kernel with smoothness parameter $\nu = 2.5$, a linear kernel, and an additive RBF–linear combination.
We also tested Automatic Relevance Determination (ARD) variants of the RBF and Matérn kernels, which allow feature-specific length scales, as well as an ARD RBF combined with a linear kernel.
For each of these kernels, we additionally considered versions that included a white noise term, resulting in 14 total configurations.

For larger datasets with at least 4000 samples, we restricted the search to the RBF-ARD plus linear kernel with an added white noise term due to computational limitations.

\subsubsection{Multilayer Perceptron (MLP)}

We conducted a grid search to optimize multi-layer perceptron (MLP) neural networks, evaluating different network architectures.
The MLPs used a consistent building block of a linear transformation followed by layer normalization, a ReLU activation, and another linear transformation, repeated across the hidden layers.

The search varied the number of hidden layers (1, 3, or 5) and the number of neurons per layer (16, 64, or 256), resulting in nine distinct architectures.
Each hidden layer had the same width within a given configuration.
The input dimension matched the dataset features, and the output dimension was one neuron for regression tasks or the number of classes for classification tasks.

Training parameters were fixed across all runs: batch size of 256, learning rate of 0.001 with the Adam optimizer~\cite{kingma2014adam}, a maximum of 1000 epochs, and early stopping.
The early stopping criterion was adaptive, with a patience of either 10 epochs or the number of input features plus one, whichever was larger.

Model selection was
performed on validation quality, using training quality as a tiebreaker when
validation quality was equal within the improvement threshold.

We conducted a grid search to evaluate a wide range of model sizes on the test results over image classification for MNIST and FashionMNIST.

\subsubsection{Convolutional (MPO)\(^2\) (C-(MPO)\(2\))}

For the C-(MPO)\(^2\) models the search instead varied the pixel rank (2, 8, or 16) and the patch rank (1, 2, or 5) at a fixed number of patches of 4 and polynomial degree 3.

\subsubsection{Convolutional Layer Network - Multilinear Perceptron (CNN-MLP)}

We use a convolutional layer, followed by an MLP layer.
The search was performed variating the number of convulation layers (1 and 3), the number of base channels (2,8,32 and 64) and the number of neurons in the hidden dimension of the MLP layer (0,16,32,128 and 256).

Code and the results of the ablation sudy for the developed procedures can be found at~\cite{myrepo} for reproducibility.

\section{Results and Discussion}
In Table~\ref{tab:regression_results} we report the \(R^2 * 100\) metric for the regression task, while in Table~\ref{tab:classification_results} we report the accuracy in percentage for classification.
The reported metrics are calculated over the test set, at the minimum validation loss epoch.
The table is organized into three groups, from top to bottom: the proposed (MPO)\(^2\) models, existing tensor decomposition based polynomial models and non-polynomial baselines for supervised learning.
The best overall model is highlighted in bold, while the best polynomial tensor network based model is underlined.

XGBoost and GP were run deterministically, and therefore, no error bars are reported for these two baselines. The missing results for GP are due to the dataset size and the computational complexity of GP as the number of features grows.
The missing results for TNML are due to training instability across all seeds for the model specifications we iterated over.
The results for TNML models flagged with \(F\) in Table~\ref{tab:regression_results} are results obtaining large negative values, and omitted for visual clarity of the table and lack of significance.

For both regression and classification, (MPO)$^2$ outperforms the other polynomial tensor network-based methods on most datasets, and when it is not the best, its performance remains close to the strongest alternative tensor based method.
Notably, it appears that feature ordering plays an important role when contrasting TNML-P and TNML-F with CPD and (MPO)$^2$.
Since CPD and (MPO)$^2$ are invariant to feature ordering, they consistently outperform the MPS/TT structures that are feature order dependent.

Comparing with standard regression and classification models (XGBoost, MLP, GP), we see that polynomial models, probably unsurprisingly, perform worse than standard deep learning methods.
This is due to the expressive capabilities of the models.
XGBoost and MLP can virtually represent any function due to the non-linearity intrinsic in the models, losing explainability and insight in how the data can produce the prediction.
In contrast, the expressive power of polynomial models is constrained by the structural choice of the degree of the polynomial they represent.
However, depending on the use case, restricting the model to a polynomial can still yield sufficiently strong performance while providing a highly detailed picture of feature interaction strengths across different orders.
This can be useful when dealing with data interpolation, where learning the parameters governing the dynamic is often the main goal as done in various applications~\cite{pr13123855, Bahmani_2024, polym13142291, Ogbo15122026, TWAROG20212014, RAI2013341}.

Gradient methods seem sufficient to efficiently learn (MPO)\(^2\) models.
In all the experiments, gradient methods are consistently faster and more memory efficient, while obtaining comparable accuracy with respect to second-order methods.
We report a time plot contrasting second order learning method with AdamW gradient descent, reporting validation accuracy during training on MNIST dataset in the supplementary material.

In the supplementary material~\cite{myrepo}, we systematically include the ablation study of the different modeling components of the (MPO)$^2$ procedure considering the Type I and Type II formulations (i.e., T1 and T2) as well as applications of the Masking (M) and Linear (L) MPOs.
From the results we observe that all the specified (MPO)$^2$ variants produce best performance within the uncertainty bounds on at least one of the considered datasets.
Consequently, the utility of the different (MPO)$^2$ variants are dataset dependent and the (MPO)$^2$ specification that is most suited for a given dataset needs to be accessed on the validation set.
The MPOs presented in this paper, namely LMPO, MMPO and CMPO, have been chosen to showcase how this framework can deal with unconstrained MPOs (LMPO), structured (MMPO) and hybrid (CMPO).
The choice of the best specification is highly dependent on prior knowledge of the specific problem and on the constrains we want to impose.
For this reason the code and the framework allow for easy implementation of specific arbitrary MPOs structures.

Note that MPOs structures can also represent or combine with CPDs structures.
MPOs that are diagonal with respect to the ranks (or that multiply element-wise) are effectively CPD-structured objects.

\begin{figure*}[t]
  \centering
  \includegraphics[width=0.98\linewidth]{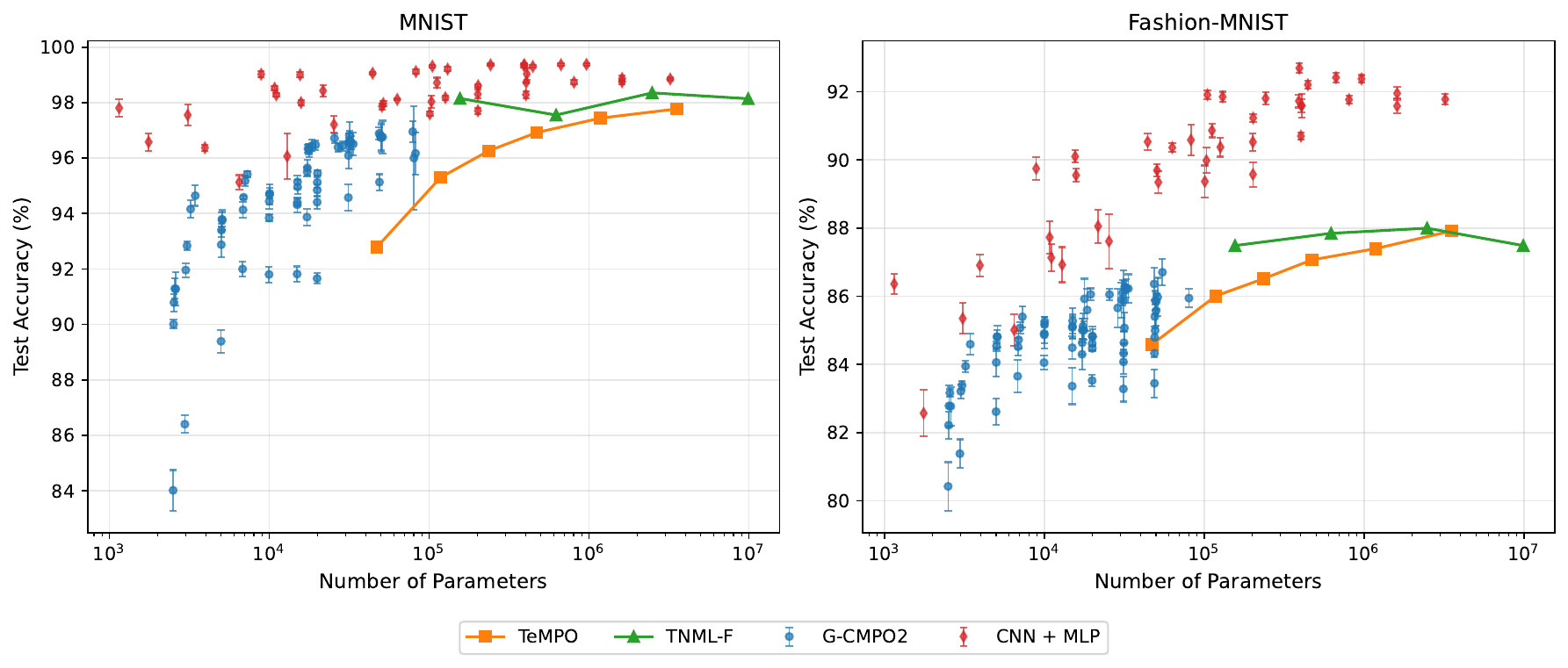}
  \caption{
    Accuracy on the test set for MNIST and Fashion MNIST classification tasks as function of parameters.
    TeMPO as described in~\cite{Ayvaz-2022-CPD-StructuredMulti} as one-vs-all classifiers for each class.
    TNML results are reported from~\cite{efthymiou2019tensornetworkmachinelearning}.
    CNN+MLP are different configuration of a Convolutional Network and an MLP.
  }\label{fig:pareto}
\end{figure*}

\subsection{Images results}\label{sec: images}

We report the average accuracy on the MNIST~\cite{deng2012mnist} and FashionMNIST~\cite{xiao2017fashion} datasets in~\Figref{fig:pareto}, as a function of the number of parameters, comparing against the CPD Type I specification as this structure was imposed for these datasets in~\cite{Ayvaz-2022-CPD-StructuredMulti}, TNML with the Fourier basis~\cite{efthymiou2019tensornetworkmachinelearning} and standard CNN-MLP classifiers.
Notably, for this image dataset we apply the Convolution MPO in our (MPO)$^2$ procedure.
Inspecting the~\Figref{fig:pareto} we observe that the (MPO)$^2$, can reach strong predictive performance using substantially fewer parameters, while reaching the same results as TNML, which due to the high number of blocks being feature dependent exhibit rapidly increasing parameters as function of ranks.

For larger image datasets like CIFAR10 and CIFAR100, second-order methods become infeasible due to the high dimensionality of the inputs.
Therefore we study the accuracy on the test set of the (C-MPO)\(^2\), learning the model parameters with gradient descent.
In the supplementary material~\cite{myrepo}, we report the test accuracy with respect to the number of parameters of the model.
The convolutional model used differ between CIFAR and MNIST datasets due to the different dimensionality of the images.
CIFAR, compared to MNIST, has an additional color channel dimension, so the patched input dimensionality is three instead of two.
This means that the independent convolution tensor network structures are three, while on MNIST only two, following the same scheme for the Convolutional (MPO)\(^2\).

The versatile specification of the multivariate polynomial by the considered (MPO)$^2$ modeling enable the systematic assessment of suitable tensor network specifications for multivariate polynomial regression with each dataset benefiting from different structures imposed.

In the supplementary material~\cite{myrepo} we present additional experiments.
We report the average test accuracy on the MNIST, FashionMNIST and CIFAR10 and CIFAR100 in function of number of parameters as well as a time analysis of one learning run, contrasting convergence speed of gradient and Newton methods for MNIST.
We explore exact polynomial inference, and devise an efficient structure identification procedure systematically growing the polynomial degree from lower degree learned (MPO)$^2$ representations that naturally avoids overfitting when considering the modeling of noise-free polynomial functions.

\section{Conclusions}
We presented the (MPO)$^2$ procedure for multivariate polynomial regression and demonstrated that this approach outperformed conventional tensor network based polynomial regression modeling procedures based on existing CPD and MPS/TT based decompositions for polynomial regression. We attribute the enhanced performance to the (MPO)$^2$ procedures to feature order independence when compared to existing MPS/TT based methodologies. Notably, we explored the versatility of the (MPO)$^2$ framework leveraging Linear, Convolutional and Masking MPO to learn compressed feature representations and accounting for weight redundancies. We expect there are many further generalizations in which the MPO formalism can be used to accommodate other types of operations. As such, we also expect the (MPO)$^2$ can be a useful tool when combined with deep learning modeling approaches akin to how the pi-sigma based CPD procedure has been imposed as nonlinear polynomial transformations of deep learning models.\\
\textbf{Limitations: }We presently only considered (MPO)$^2$ modeling procedures in which the rank was specified to be identical across the MPO blocks. Future work should consider how individual ranks can be efficiently learned which would require an exponential evaluation of model specifications. It should also explore how Bayesian inference procedures can be used to quantify parameter uncertainty and automatically learn the relevance of different rank terms, see also~\cite{Hinrich_2020,kilic2025interpretablebayesiantensornetwork}.
Pure end-to-end polynomial models cannot outperform non-linear deep learning models in regression and classification, due to the limited representation power of the function space.
As seen in polynomial models like $\Pi$‑nets~\cite{Chrysos2021}, for image classification the results are highly improved with the introduction of a non-linearity. We do not use non-linear transformations to represent the parameters of the polynomial, while non-linear interactions in the features could greatly improve the results.

All the code to reproduce the results can be found at Repository~\cite{myrepo}.

\paragraph*{Acknowledgements:} This project was supported by the Novo Nordisk Foundation, grant no. NNF23OC0083524.

\bibliographystyle{IEEEtran}
\bibliography{bibliography}
\end{document}